%% file: main.tex
\def\year{2022}\relax
\documentclass[letterpaper]{article} % DO NOT CHANGE THIS
\usepackage{aaai22}  % DO NOT CHANGE THIS
\usepackage{times}  % DO NOT CHANGE THIS
\usepackage{helvet}  % DO NOT CHANGE THIS
\usepackage{courier}  % DO NOT CHANGE THIS
\usepackage[hyphens]{url}  % DO NOT CHANGE THIS
\usepackage{graphicx} % DO NOT CHANGE THIS
\usepackage{natbib}  % DO NOT CHANGE THIS AND DO NOT ADD ANY OPTIONS TO IT
\usepackage{caption} % DO NOT CHANGE THIS AND DO NOT ADD ANY OPTIONS TO IT
\DeclareCaptionStyle{ruled}{labelfont=normalfont,labelsep=colon,strut=off} % DO NOT CHANGE THIS
\usepackage{algorithm}
\usepackage[noend]{algorithmic}
\usepackage{def}
\usepackage{newfloat}
\usepackage{listings}
\floatstyle{ruled}
\newfloat{listing}{tb}{lst}{}
\floatname{listing}{Listing}
\title{Reversible Action Design for Combinatorial Optimization with Reinforcement Learning}
\author {
    Fan Yao\textsuperscript{\rm 1},
    Renqin Cai\textsuperscript{\rm 1},
    Hongning Wang\textsuperscript{\rm 1}
}
\affiliations {
    \textsuperscript{\rm 1} Department of Computer Science, University of Virginia, USA\\
    fy4bc@virginia.edu, rc7ne@virginia.edu, hw5x@virginia.edu,
}

\usepackage{bibentry}
\begin{document}

\maketitle
% \frenchspacing

\input{abstract}
\input{intro}
\input{related}

\input{method}

\input{exp}
% \input{conclusion}
% \bibliography{ref}
% \bibliographystyle{unsrtnat}
\twocolumn
\bibliography{ref}
% \newpage
% \input{appendix}

\end{document}

%% file: abstract.tex
 \begin{abstract}
Combinatorial optimization problem (COP) over graphs is a fundamental challenge in optimization. Reinforcement learning (RL) has recently emerged as a new framework to tackle these problems and has demonstrated promising results. However, most RL solutions employ a greedy manner to construct the solution incrementally, thus inevitably pose unnecessary dependency on action sequences and need a lot of problem-specific designs. We propose a general RL framework that not only exhibits state-of-the-art empirical performance but also generalizes to a variety class of COPs. Specifically, we define state as a solution to a problem instance and action as a perturbation to this solution. We utilize graph neural networks (GNN) to extract latent representations for given problem instances for state-action encoding, and then apply deep Q-learning to obtain a policy that gradually refines the solution by flipping or swapping vertex labels. Experiments are conducted on Maximum $k$-Cut and Traveling Salesman Problem and performance improvement is achieved against a set of learning-based and heuristic baselines.

\end{abstract}

%% file: intro.tex
\section{Introduction}

Combinatorial optimization problems (COP) have attracted extensive interest from both machine learning and operation research communities in decades because of its pervasive application scenarios \cite{grotschel1991optimal, plante1987product, waldspurger2015phase, candes2015phase}. Most of these problems are known to be NP-hard \cite{karp1972reducibility} and extremely challenging for their combinatorial nature: the optimization is performed over a discrete structure (e.g. a weighted graph) and is often associated with exponentially sized feasible solution space. To give a concrete example, consider Maximum $k$-Cut: given a weighted graph $G=(V, E)$, find a partition of $V$ that divides $G$ into $k$ disjoint sets which maximize the sum of edge weights among each pair of partitions. Another popular example is Traveling Salesman Problem (TSP), in which the goal is to search for the shortest possible route that visits each node on a graph once and only once and returns to the origin node. 

Since the solution domain of COP is prohibitively large, exact methods, such as enumeration based approaches are simply intractable. Therefore, classical solutions for COPs mainly focused on heuristics for decades; such methods include simulated annealing \cite{patel1991preliminary} and genetic algorithms \cite{dedieu2003design}.
%, which all have proven to be acceptable in many practical applications \cite{sorensen2013metaheuristics}.
However, these heuristics often suffer from arduous case-specific design and redundant computation because common combinatorial structures have to be addressed repeatedly across similar problem instances. Recently, learning-based methods have emerged as an effective tool to provide generalizability across similar problems by exploiting the structure of the target problem. In particular, reinforcement learning (RL) has been identified as a powerful end-to-end framework \cite{bello2016neural, ma2019combinatorial,kool2018attention,abe2019solving,ecodqn, huai2020malicious}, because it does not rely on ground-truth solutions compared to supervised methods \cite{vinyals2015pointer, li2018combinatorial, mittal2019learning, nowak2017note, joshi2019efficient,sarker2020deep,wu2020deja}. However, most RL-based attempts are tailored for a single class of COPs and require a lot of domain-specific designs. For instance, \citet{bello2016neural} adopt the Pointer Network structure \cite{vinyals2015pointer} to encode a permutation of vertex set for TSP, but it is specifically designed to tackle COPs for which the output depends on the length of the input. \citet{ma2019combinatorial} resort to a hierarchical RL structure to address TSP with time-window constraints, but such design is only tailored for TSP and can hardly generalize to other COPs. To alleviate the issue, \citet{khalil2017learning} proposed a general framework, S2V-DQN, to cope with a wide range of COPs by combining graph neural network (GNN) and deep Q-learning. However, S2V-DQN's node selection strategy is greedy in nature, which prevents it from yielding high-quality solutions since the learned policy cannot revoke its previous decisions based on the observations afterward. Moreover, this greedy node selection strategy suffers when the target COPs have non-sequential structured solutions, e.g., Maximum $k$-Cut with $k>2$, because a sequence of irreversible node selections is not an effective way to form graph partitions. 
 
We propose a new general RL framework for COPs with reversible action design. Similar to S2V-DQN, our method also consists of a GNN encoder for state-action encoding and a deep Q-network for policy learning. However, unlike S2V-DQN and other similar approaches in which the agent builds solutions with irreversible actions, our formulation allows RL agents to explore the whole solution space starting from an arbitrary solution and keep improving it. Another strength of our framework is the ability to handle a wider range of COPs with different solution structures. 
%Compared to ECO-DQN \cite{ecodqn}, which only handles Maximum 2-Cut problem, 
Our method precludes any ad-hoc designs in the RL formulation and leaves problem-specific properties to the state-action encoding networks. As a result, it can easily transfer to different COPs by applying different state-action encoders while maintaining all other essential designs in the RL module. We tested our framework on two extensively studied COPs, Maximum $k$-Cut and TSP, on both synthetic and real-world datasets and obtained competitive performance compared to both learning-based methods and heuristic baselines. In addition to the encouraging performance, we also observed that the trained agent has learned to approach the optimal solution with the ability to jump out of the local minima along the way, which is the key bottleneck of greedy or heuristic based solutions.

% In the following sections, we will first introduce the background related to our proposed framework and formalize our method in detail. Then we'll apply our framework on two extensively studied COPs, Maximum $k$-Cut and TSP, and further present experimental results on several synthesized datasets along with the comparison with S2V-DQN, ECO-DQN, and other traditional approaches. Finally, we'll discuss some fundamental challenges that we identified in our experiments. We believe that our work could shed some light on the algorithm design for the combinatorial optimization community. 

%% file: related.tex
\section{Related Works}

The recent success of applying machine learning to solve COPs can be traced back to \citet{vinyals2015pointer}, where pointer networks are used to solve TSP via supervised learning. Then \citet{bello2016neural} further developed an RL solution for TSP based on pointer networks using policy gradient. After that, a number of follow-up works emerged to focus on various COPs with different challenges. For example, \citet{kool2018attention} applied the Transformer architecture as the state encoder and reported state-of-the-art results on vehicle routing problems. \citet{ma2019combinatorial} resorted to a hierarchical RL structure to address TSP with time-window constraints. And \citet{abe2019solving} trained a GCN with Monte-Carlo tree search to enhance its generalizability on five NP-hard COPs. Besides these attempts, S2V-DQN \cite{khalil2017learning}, to the best of our knowledge, is the first general RL-based solution for COPs. It combines a graph neural network (GNN) and deep Q-network and is demonstrated to be effective across three classes of COPs (i.e. Minimum Vertex Cover, TSP, and Maximum 2-Cut). 

All the aforementioned RL-based methods employ a greedy node selection strategy to construct solutions, i.e., classifying one unlabeled node at a time until all nodes are labeled. Instead of constructing the solution in an incremental manner, \citet{ecodqn} proposed ECO-DQN for Max Cut, an exploratory method that allows the RL agent to continuously improve any given solution via local search \cite{johnson1988easy}, by modeling the actions as vertex-flipping. This distinctive reversible action design enables the agent to access any valid solution during planning. The reported empirical improvement from this reversible action design suggests its advantage over its irreversible counterpart.  
%They named the RL agents in ECO-DQN and those in former RL formulations as reversible/irreversible ones accordingly and claimed a remarkable improvement of reversible agents compared to irreversible ones. 
However, ECO-DQN involves several ad-hoc designs in its modeling pipeline (such as reward shaping and input feature engineering), which makes it less applicable to other COPs. 

% However, the generalizability of ECO-DQN seems to be limited due to two observations. First, the vertex-flipping action design only applies to COPs with simple solution structures, e.g. Maximum 2-Cut. For problems like Maximum $k$-Cut ($k>2$) and TSP, the algorithm design pattern cannot trivially generalize.

% Apart from value-based and policy-based RL, another direction is applying GNN with a tree search: \cite{li2018combinatorial} utilized a GCN along with a guided tree-search in a supervised manner on four canonical NP-hard problems, and , \cite{abe2019solving} trained a GCN with Monte-Carlo tree search as a boosting module.

%% file: method.tex
\section{Method}
\subsection{Problem Formulation}
%In a nutshell, we formulate the common definition of combinatorial optimization problem (COP) as follows: $S$ is a finite set of elements, $f: S \xrightarrow[]{} \mathbb{R}$ is a cost function, and the aim of a COP is to find the optimal value of $f$ on $S$. Since the domain $S$ is finite, the global optimum always exists. But in most cases, $S$ is formidably large, and therefore trivial enumeration-based solutions are not feasible. In this work, we focus on the COPs where $S$ displays a graph structure. With a slight abuse of terminology, we use COP to denote combinatorial optimization problems on graphs. In this case, we can further specify the essential components of a COP as follows:

In this work, we focus on the COPs where the problem instances impose a weighted graph structure. With a slight abuse of terminology, we use COP to denote combinatorial optimization problems on graphs. In this case, we can specify the essential components of a COP as follows:
\begin{enumerate}[leftmargin=*]
    \item A problem instance, i.e. a weighted graph $G\!=\!(V,\! E,\! w)$ sampled from some underlying distribution $\mathcal{D}$, where $V\!=\!(v_1,\cdots,v_n)$, $E\!=\!(e_1,\cdots,e_m)$ and $w\!=\!\{w(v_i, v_j)\}$ represent the vertex set, the edge set and the weights associated with each edge, respectively. For simplicity, we denote $w_{ij}\!=\!w(v_i, v_j)$ and simply assign $w_{ij}\!=\!0$, if $v_i$ and $v_j$ are not directly connected. In this case, $G$ can be fully expressed by the weighted adjacency matrix $w$.
    \item A solution to a problem instance $G$ given by a mapping $l$ from $V$ to $\mathbb{N}$, which assigns an integer-valued label to each vertex. For simplicity, we denote the solution as $L\!=\!(l(v_1), \cdots, l(v_n))$.
    \item An objective function $O(G, L) \!\in \!\mathbb{R}$ that takes a problem instance $G$ and a solution $L$ as input and outputs the evaluation on $(G, L)$. The goal of a COP is to find a global optimal solution $L^*_G\!=\!\text{argmin}_{L}O(G, L)$ under certain constraints to $L$.
\end{enumerate}
Since the solution space on a finite graph $G$ is also finite, its global optimum $L^*_G$ always exists if the problem is feasible. But in most cases, as the solution space of $l$ is formidably large, trivial enumeration-based solutions are not feasible. 

Given the abstract formulation above, we can instantiate some popular COPs by specifying different $G$, $L$ and $O$. For example, consider Maximum $k$-Cut, where $G$ is an arbitrary weighted graph, $L$ is an element from $\mathcal{L}\!=\!\{(l_1, \cdots, l_n)| l_i \!\in\! \{1, 2, \cdots, k\} \}$, and $O\!=\!\sum_{c=1}^{k}\sum_{(i,j) \in \{i,j|l_i=l_j=c\}}w_{ij}$. Similarly we can formulate TSP as: $G$ is a complete graph characterized by a weighted adjacency matrix $w$, $L$ is an element in the set of all permutations of $\{1,2,\cdots,n\}$, and  $O\!=\!w_{l_1l_n}\!+\!\sum_{k=1}^{n-1}w_{l_kl_{k+1}}$.
In this paper, we illustrate our solution framework using Maximum $k$-Cut and TSP, because they represent two typical classes of COPs in a broad sense. TSP represents COPs with sequentially structured solutions, such as vehicle routing problem \cite{dantzig1959truck}; and Maximum $k$-Cut represents COPs that fall into the category of node classification, such as Minimum Vertex Cover and Maximal Independent Set. By designing meta-algorithms for solving Maximum $k$-Cut and TSP, we demonstrate how our methodology can shed light on finding solutions for a wide range of related COPs.
\subsection{Reinforcement Learning Solution Design}

We apply reinforcement learning to design a meta-algorithm that minimizes the discrete objective function $O(G, L)$. Unlike the greedy strategy employed by S2V-DQN \cite{khalil2017learning}, in which the agent is trained to construct the solution sequentially by adding nodes to a partial solution, our design allows the agent to gradually improve a given solution $L$ by adding small perturbations to it at each step. Specifically, we define the primitive components in our RL formulation as follows:
\begin{enumerate}[leftmargin=*]
\item State space. The state space $\mathcal{S}\!=\!\{s\!=\!(G, L)|G \in \mathcal{D}, L\in \mathcal{L}_G\}$ ($L$ may depend on $G$) is defined as the set of all the valid graph-solution pairs. Note that the distribution $\mathcal{D}$ might contain graphs with different sizes. 
\item Action space. The action space $\mathcal{A}(s)$ is designed as all valid perturbations on $L$ given a state $s\!=\!(G, L)$, plus a dummy action $\emptyset$ which terminates further actions. $\mathcal{A}(s)$ includes those perturbations such that the perturbed $L$ is still a valid solution in $\mathcal{L}$ for $G$. In particular, we define two types of perturbations on $L$:
\begin{enumerate}
\item  Flipping $\mathcal{A}^{flip}$, in which an action $a$ flips the label $l_i$ for node $v_i$ into a different one.
\item  Swapping $\mathcal{A}^{swap}$, in which an action $a$ swaps the label $l_i$ and $l_j$ for node pair $(v_i, v_j)$.
\end{enumerate}
\item Reward. The reward function $r(s, a)$ is defined as the negative value change for the objective function $O$ after taking action $a$ at state $s\!=\!(G, L)$, i.e., $r(s, a)\! =\! O(G, L) \!- \!O(G, a \circ L)$.
\item Policy. We adopt the Q-learning framework to fit a Q-function $Q(s, a)$ that evaluates the accumulated reward given any state-action pair. According to the learned q-value, we apply a deterministic policy $\pi(s)\!=\!\argmax_{a \in \mathcal{A}(s)} Q(s, a)$.
\item Termination. An episode is terminated when the agent decides to take the dummy action $\emptyset$, or the maximum number of actions $M$ have been executed, where $M$ is a hyper-parameter and grows with the graph size $n$.
\end{enumerate}
Apart from the difference in action design, another distinction between our work and S2V-DQN lies in the construction of states: in S2V-DQN, the state is a graph with a partial solution; while in our framework the state is a graph with a complete solution. As a result, our solution can always exploit the complete structure of a solution at each step, while S2V-DQN struggles with incomplete information to take the next action, especially in the first few steps when the partial solution only consists of a few scattered nodes. Another advantage of our framework is its flexibility when dealing with inherent or additional constraints, e.g., the permutation constraint in TSP (i.e. no duplicated nodes on a route), or the cut-size constraints in Maximum $k$-Cut (i.e. given size of each subgraph induced by the cut). To address these constraints, we only need to pose restrictions on the action space to guarantee that each perturbed solution also satisfies the constraints. But in S2V-DQN and other works, such as \cite{bello2016neural} which also adopts S2V-DQN's greedy design pattern, ad-hoc changes to the architecture of the Q-network have to be made to handle the constraints.

\subsection{State/Action Representation and the Q-function}
To apply Q-learning for solving COPs, we need to obtain a continuous representation of any state-action pair $(s, a)$. This representation should incorporate both the combinatorial nature of graph $G$ and the structure of a solution $L$. Considering these factors, we adopt the message passing neural networks (MPNN) architecture \cite{gilmer2017neural}, which is a general graph neural network framework to obtain node representations by collecting information iteratively from local graph structures. In detail, we first initialize an embedding vector $\mu_v^0=\bm{0}\in\mathbb{R}^{d}$ for each node $v$; then at round $k$, $\{\mu_v^k\}$ is updated by leveraging information from its neighbors by:
% \begin{tiny}
\begin{equation}\label{eq:graph-encoder}
\begin{aligned}
    \mu_v^{k+1}=\text{relu}\Big(\theta_0 x_v + \theta_1 \frac{1}{|N(v)|}\sum_{u\in N(v)}w_{uv}\mu_u^k \\+ \theta_2 \frac{1}{|N(v)|}\sum_{u\in N(v)}\text{relu}(\theta_3 w_{uv})\Big),
\end{aligned}
\end{equation}
% \end{tiny}
where $\theta_0, \theta_2 \in \mathbb{R}^{d\times p}, \theta_1 \in \mathbb{R}^{d\times d}$ and $\theta_3 \in \mathbb{R}^p$ are model parameters, $N(v)$ is the set of node $v$'s neighbors and $|N(v)|$ is its cardinality. In the first term, $x_v \in \mathbb{R}^p$ is the static feature vector for vertex $v$ that incorporates additional node information from a problem instance. For instance, in Maximum $k$-Cut, $x_v$ can take the form of a one-hot label vector of $v$ to encode the current cut information; and in TSP, $x_v$ can be the initial coordinates of $v$ if given. The second term in Eq \eqref{eq:graph-encoder} aims to aggregate information from $v$'s neighbors by taking a weighted average over the neighbors' embeddings proportional to the edge-weights. This encodes neighborhood information together with their edge weight defined relatedness. And the third term is served to emphasize if $v$ is closely connected to its neighbors based on the given edge weights. 
%Compared to the GNN structure employed in S2V-DQN, there are two main differences in our design: First. we replaced the vanilla average in the second term with a weighted one, which better captures the local structure; 2. we scale the second and the third term with a constant factor $\frac{1}{|N(v)|}$, which, based on our observation, plays an essential role in the generalization to larger problem instances. 

The final node embeddings $\{\mu_v^{T}\}$ are obtained by iterating this procedure for $T$ rounds, such that they are expected to carry $T$-hop information among the nodes based on the graph topology. The state and action representations can thus be constructed based on the computed node embeddings $\{\mu_v^{T}\}$. For different COPs, we need different readout functions to compute the graph embedding from node embeddings. Take Maximum $k$-Cut and TSP as two examples to illustrate this process.

\noindent{\textbf{$\bullet$ Maximum $k$-Cut}}, suppose the current graph cut given by state $s=(G,L)$ is $(V_1, \cdots, V_k)$, where each $V_i$ corresponds to a cluster of nodes. We first define the cluster representation by averaging all the node vectors within the cluster as $H_c(i)=\frac{1}{|V_i|}\sum_{v \in V_i} \mu_v^{T}$. For the action embedding, we represent a flipping action (i.e., to flip $u$'s label from $i$ to $j$) as the concatenation $H_a^{flip}=[\mu_u^{T}; H_c(j)]$, where $\mu_u^{T}$ is the representation of vertex $u$, which is expected to carry information about vertex $u$ and cluster $i$, and $H_c(j)$ is the representation of the target cluster $j$. Similarly, we can also define the representation of the swapping action (i.e., to swap the labels of $u$ and $v$) as $H_a=[\mu_u^{T}; \mu_v^{T}; H_c(l(u)); H_c(l(v))]$, where $l(u)$ and $l(v)$ correspond to $u$ and $v$'s current labels. Finally, we use attention to construct the state representation over cluster embeddings. The attention weight is computed by taking cluster embeddings as the reference and the action embedding as the query: $w(i) = \text{softmax}_i(H_c^\top(i)W_a H_a)$,
where $H_c(i) \in \mathbb{R}^{d}, H_a \in \mathbb{R}^{4d}$, and $W_a \in \mathbb{R}^{d\times 4d}$ is a trainable parameter matrix. Then the state representation is formed by $H_s=\sum_{i=1}^{k} w(i)H_c(i)$.

\noindent{\textbf{$\bullet$ TSP}}, its solution exhibits a sequential structure instead of a clustered one. Therefore, we adopt an RNN encoder to get the state embedding: $H_s = RNN(\{\mu^T_i\}_{i=1}^n;\Theta^{RNN})$, where $\{\mu^T_i\}_{i=1}^n$ is the sequential input of RNN defined by the node permutation in solution $L$, $\Theta^{RNN}$ is the weights of RNN, and the output $H_s$ is taken as the RNN hidden state at step $n$. Because of the constraint in TSP, i.e., no repeated nodes on a tour, flipping action does not apply. To specify the swapping action, we define the sequential-swap $(i, j), i<j$, for TSP as to swap the sub-sequence $(L_i, L_{i+1}, \cdots, L_j)$ in $L$ to $(L_j, L_{j-1}, \cdots, L_i)$. Because a sequential-swap action $(i, j)$ only changes edge-weights associated with $v_{i-1},v_i, v_j,v_{j+1}$ in the TSP tour (i.e., change edges from $v_{i-1,i}, v_{j,j+1}$ to $v_{i-1,j}, v_{i,j+1}$), we represent it by $H_a=[\mu_i^{T};  \mu_j^{T}; \mu_{i-1}^{T}; \mu_{j+1}^{T}]$, where $1\leq i<j\leq n, \mu^T_0=\mu^T_n, \mu^T_{n+1}=\mu^T_1$. 

Based on the state and action representations, we establish the parameterized Q-function as $Q(s, a)=W_0 \cdot \text{relu}([W_1H_s; W_2H_a])$, where $W_0 \in \mathbb{R}^{1 \times 2d}, W_1\in \mathbb{R}^{d\times d_s}, W_2\in \mathbb{R}^{d\times d_a}$ are trainable parameters, $d_s$ and $d_a$ are the dimensions of state and action embeddings respectively. We use swap action design in Maximum $k$-Cut as an example in Algorithm \ref{q-learn} to illustrate the procedure of applying our proposed RL solution framework for COPs.

\subsection{End-to-End Training}\label{end2end}
We apply $N$-step off-policy TD method \cite{sutton2018reinforcement}, i.e., $N$-step Q-learning to train $Q(s, a)$, which has been demonstrated to be effective when dealing with delayed rewards. By evaluating the value function $N$-step ahead, we encourage the agent to be less myopic by avoiding eagerly punishing an action which induces a negative immediate reward. At each training step, we first sample a random batch of state-action-reward tuples $\mathcal{B}\!=\!\big\{(s^0_1, \{a^j_1\}_{j=1}^N, \{r_1^j\}_{j=1}^N, s^N_1), \cdots, (s^0_b, \{a_b^j\}_{j=1}^N, \{r_b^j\}_{j=1}^N, s^N_b)\big\}$ from the replay buffer $\mathcal{M}$, where the index $j=1,\cdots,N$ denotes the observations at the $j$-th step from the corresponding state $s^0$. The q-loss is given by $\sum_{i=1}^b\big(y_i\!-\!Q(s_i, a_i; \Theta)\big)^2$, where the target $y_i$ is computed as the accumulated reward in $N$ steps starting from $s^0_i$ and plus the estimated long-term rewards starting from $s_i^N$:
\begin{equation}\label{q-tar}
    y_i = \sum_{j=1}^N \gamma^{j-1} r_i^j + \gamma^N \argmax_a Q(s_i^N,a;\hat{\Theta}),
\end{equation}
where $\gamma\!\in \!(0, 1)$ is the discounting factor and $\hat{\Theta}$ is the currently estimated parameter in the target Q-net.

%\begin{figure}
%\centering
%\includegraphics[width=1.0\textwidth]{resource/dqn_illus.png}
%\caption{\label{fig:dqn_struct}The DQN architecture for Maximum $k$-Cut}
%\end{figure}
\begin{algorithm}[t]
   \caption{Q-learning for swap search}
   \label{q-learn}
\begin{algorithmic}
   \STATE {\bfseries Input:} Training epochs $E$, episode maximum length $M$, exploration constant $\varepsilon$. 
%   \REPEAT
\STATE {\bfseries Initialization:} Set experience replay buffer $\mathcal{M}=\emptyset$.
    \FOR{$epoch \leftarrow 1$ to $E$} 
    \STATE Sample a graph $G(V, w)$ from $\mathcal{D}$.
    \STATE Initialize a random solution $L_0$. 
    \FOR{$t \leftarrow 0$ to $T$}
        \STATE with probability $\varepsilon$, select a random node pair $a_t = (v_i, v_j)$; othereise, $a_t=\text{argmax}_{(i,j)}Q(s_t, (v_i, v_j);\Theta)$.
        \STATE Swap the $i$-th and $j$-th elements in $L_t$ to give $L_{t+1}$, assign $s_{t+1}=(G, L_{t+1})$.
        \STATE Execute the evaluation function $O(G, L_{t+1})$ to observe reward $r_t$.
        \IF {$t \geq N$} 
            \STATE Add tuple $(s_{t-N}, a_{t-N}, r_{t-N,t}, s_{t})$ to $\mathcal{M}$.
            \STATE Sample a batch $\mathcal{B}$ from $\mathcal{M}$.
            \STATE Update $\Theta$ for $\mathcal{B}$ via gradient descent.
        \ENDIF
    \ENDFOR
\ENDFOR
   \STATE {\bfseries Output:} Model parameter $\Theta$.
\end{algorithmic}
\end{algorithm}

The main challenge for Algorithm \ref{q-learn} lies in the evaluation of optimal action in planning: computing the maximum of $Q(s, a)$ over all actions in Eq \eqref{q-tar} could be extremely expensive, especially when we adopt the swapping action space $\mathcal{A}^{swap}$ design, which is of size $O(n^2)$ and $n$ is the number of nodes. In this case, the time and space complexity for a single forward operation is $O(n^2d^2)$ and $O(n^2d)$ respectively, where $d$ is the hidden dimension of state representation. Inspired by \cite{van2020q,dulac2015deep}, we introduce an auxiliary action-proposal network to perform action elimination. The idea is that after the state embedding $H_s$ is obtained for the current state $s$, we directly propose a pseudo action $\Tilde{a}(s)$ in $\mathbb{R}^d$ from the auxiliary network: $\Tilde{a}(s) \!= \!AP(H_s;\Theta_{ap})$.

Although $\Tilde{a}(s)$ might not directly map to a real action, $\Tilde{a}(s)$ is expected to lie around the optimal action in state $s$. Therefore, we can generate a distribution $\pi^{prop}(a)$ over the whole action space 
%$\mathcal{A}^{swap}=\{a=(v_i, v_j)| i<j, l_i\neq l_j \}$ 
by leveraging the similarity between $\Tilde{a}(s)$ and each node's embedding $\mu^T_v$. For example, for swapping actions, we set $\pi^{prop}(v_i, v_j)\! \propto \!\exp\big(s(v_i)\! +\! s(v_j)\! + \!\theta_0 s(v_i) s(v_j)\big)$, where $s(v) \!=\! \Tilde{a}(s)^\top \mu^T_v$ and $\theta_0\! \in\! \mathbb{R}$ is a trainable parameter. 
To perform action elimination, we first choose an action reserve ratio $\epsilon \!\in\! (0, 1]$, then draw a subset $\mathcal{A}^{swap}_{prop}$ of size $\epsilon|\mathcal{A}^{swap}| \!\propto \!O(n)$ from $\mathcal{A}^{swap}$ according to $\pi^{prop}(a)$, and restrict the search space of actions in Eq \eqref{q-tar} to $\mathcal{A}^{swap}_{prop}$. By deploying the auxiliary action proposal network, we reduce the time and space complexity for a single forward operation to $O(n^2\!+\!nd^2)$ and $O(n^2\!+\!nd)$. In practice, we observe that the hidden dimension $d$ should always increase proportionally to $n$ in order to guarantee good empirical performance. And this action elimination technique is essential to make the RL framework applicable to large-scale COPs.

To train the action proposal network, we introduce a regularized loss function:
\begin{equation} \label{prop_loss}
    \mathcal{L}(\Theta_{ap}; s)=-\log \pi^{prop}(a^*(s)|\Theta_{ap}) -\lambda H(\pi^{prop}(a|\Theta_{ap})),
\end{equation}
where $a^*(s) = \argmax_{a \in \mathcal{A}^{swap}_{prop}} Q(s,a;\hat{\Theta})$ is the action selected by the Q-net among the proposed action set $\mathcal{A}^{swap}_{prop}$. The first term in Eq \eqref{prop_loss} is to minimize the negative log-likelihood for the selected action, which makes the action with the highest Q-value more likely to be proposed. The second term is the negative entropy of the proposal distribution, and minimizing it would encourage uncertainty in $\pi^{prop}$ throughout training and prevents it from collapsing to a deterministic distribution.

%\begin{table}
%\centering
%\begin{tabular}{c|c|c}
%cost & time & space \\\hline
%Without A.P. Net.& $O(n^2d^2)$ & $O(n^2d)$\\
%With A.P. Net.& $O(n^2 + nd^2)$ & $O(n^2 + nd)$
%\end{tabular}
%\caption{\label{tab:time reduce}Time and space complexity reduction after applying action proposal network.}
%\end{table}

% \subsection{Active Searching}

%% file: exp.tex
\section{Experiments}

\paragraph{Problem Setups}
We evaluate our proposed solution framework on Maximum $k$-Cut and TSP. For Maximum $k$-Cut, we consider two settings: Maximum Cut (i.e., $k$=2) and Maximum $k$-Cut with size constraint (i.e., $k>2$ and the sizes of the graph cut are given). For the training, validation and test set of Maximum Cut, we use complete graphs whose weighted adjacency matrices are given by the pairwise Euclidean distances among $n$ nodes uniformly sampled from $[0, 1]^h$. For Maximum $k$-Cut, we use a synthetic dataset generated by $k$-clustered graph, where each problem instance $G=(V, E, w)$ is a complete graph generated by first sampling $k$ centroids $\{c_1, \cdots, c_k\}$ uniformly from $[0, 1]^h$, and then sample $m$ nodes $\{x_{ij}\}_{j=1}^{m}$ for each centroid $c_i$ from Gaussian distribution $N(c_i, \sigma^2_i I_h)$, where $m, h, \sigma_i$ are hyper-parameters and $I_h$ is the identity matrix of size $h$. The weighted adjacency matrix $w$ is then computed as the Euclidean distances between each pair of nodes, i.e., $w_{ij}=||x_i-x_j||_2$. For TSP, we draw training graphs from a distribution where each node $x_i$ is uniformly sampled from a 2-D square $\{(x, y)| 0\leq x \leq 1, 0\leq y \leq 1\}$ and the weighted adjacency matrix $w$ is computed by Euclidean distances accordingly, i.e., $w_{ij}=||x_i-x_j||_2$. For the testing set of TSP, we resort to a public benchmark TSPLIB derived from real-world instances \cite{TSPLIB}. 

Note that in our settings all training and validation graphs are complete graphs, which is computationally cumbersome when $n$ becomes large. To scale up for larger $n$ (e.g., $n>50$), we use the $K$-nearest neighbor graph ($K = 50$) to replace the complete graph, i.e., reserve top-$K$ nearest neighbors in the weighted adjacency matrix. We used two types of synthetic graphs and a real-world dataset from TSPLIB \cite{TSPLIB} in the experiment. For more detail of our configuration and the hyper-parameter setting, please refer to our full arxiv version \url{https://arxiv.org/pdf/2102.07210.pdf}.

\paragraph{Metrics and baselines}

For Maximum Cut, we consider two RL-based baselines S2V-DQN \cite{khalil2017learning} and ECO-DQN \cite{ecodqn}. In addition, we consider several popular heuristic-based methods including semidefinite programming (SDP) \cite{sdp}, genetic programming (GP) \cite{gp}, and greedy algorithm. SDP is a non-greedy algorithm that gives a solution by relaxing the discrete problem to a continuous one and then applying semidefinite programming to address the resulting optimization problem. GP is an evolutionary computation method that imitates biological evolution by iteratively improving a set of solutions through mutations and selections. The greedy algorithm we use here is an iterative method that starts with a random cut and flips or swaps the label of a vertex at each step with the greatest immediate increase in the cut value until no further improvement can be made, which corresponds to a RL agent in our framework that always follows the one-step reward.  

For Maximum $k$-Cut ($k>2$) with size constraint, S2V-DQN and ECO-DQN cannot directly apply. As a result, we use three heuristic baselines, i.e., SDP, GP, greedy algorithm for benchmarking. Note that the greedy algorithm we use here takes swapping as an action to guarantee the size constraints. 

For the baselines of TSP, we include S2V-DQN and two approximation algorithms Farthest insertion (Farthest) and 2-opt, which are reported as the two best-performing approximation methods on TSPLIB dataset \cite{khalil2017learning}. The implementation details for these two approximation algorithms can be found in \cite{applegate2006traveling}. Note that the 2-opt method is essentially the greedy algorithm that takes the swapping action as we designed for TSP.

We use the approximation ratio $R^{app}=O^*/O^{opt}$ as a metric to evaluate the quality of solutions, where $O^*$ is the best objective value given by an algorithm, and $O^{opt}$ is the true optimal objective value. For Maximum $k$-Cut problems with large sizes, we do not have access to $O^{opt}$. Considering that the greedy algorithm demonstrates itself as the strongest baseline for large-scale Maximum $k$-Cut problems, we apply it multiple times with different initial solutions and take its best result as the alternative for $O^{opt}$. For the test problem instances of TSP, $O^{opt}$ is provided in the TSPLIB dataset.

\paragraph{Solution quality comparison}

To compare the quality of solutions, we train our algorithm with problem instances up to 200 nodes and test it on 100 held-out graphs of the same size. The maximum episode length is set to twice the size of the graph; and during each training epoch, the training batch is sampled uniformly from $\mathcal{D}$ with a randomly initialized solution, with the batch sizes ranging from 50 to 500 to meet the memory limitation posed by a single graphic card. The agent starts planning from a randomly initialized solution on each test instance. Table \ref{tab:kcut} and \ref{tab:tsp} summarize the results on the average approximation ratio across Maximum Cut, Maximum $k$-Cut, and TSP with variance over multiple trials (i.e. different trials are the trajectories starting from randomly drawn initial states) shown on the superscripts and subscripts. Our algorithm is denoted by the name LS-DQN (i.e. Local Search DQN). 

For Maximum Cut with $k=2$, the flipping action design is employed in our solution. As shown in Table \ref{tab:kcut}, LS-DQN performed as good as ECO-DQN and outperformed S2V-DQN on larger test graphs. Compared to approximation heuristics such as SDP and GP, LS-DQN showed a better and more stable performance, and the gain became more significant as the test graph size increases. It is not a surprise to observe the greedy baseline has strong performance, because it can always guarantee a local minimum (i.e., there is no flipping that generates immediate improvement); and in Maximum Cut, local minima were close to the global minimum with very high probability based on our observations. The performance gain of LS-DQN mainly comes from its ability to jump out of those local minima during planning. For Maximum $k$-Cut, we test our model on three different graph scales: $(k, m)=(5,6), (10,10), (10,20)$. As shown in the bottom of Table \ref{tab:kcut}, LS-DQN performed significantly better than SDP and GP, and slightly better than the greedy algorithm. Table \ref{tab:tsp} shows the result for TSP. The Farthest and 2-opt algorithms are the two strongest baselines for the TSPLIB dataset according to the results reported in \cite{khalil2017learning}. As the graph sizes in TSPLIB are not identical to our training size, we train our RL agent on graphs with fixed sizes and test it on-the-fly on a batch of graphs whose sizes fall in a certain range, and report the result on the best tour encountered over the training epochs, like it was done in \cite{khalil2017learning}. Our LS-DQN reaches the best approximation ratio on graph sizes up to 200 compared to S2V-DQN and the other two heuristics.

\paragraph{Generalization to larger graphs}
To investigate the generalization ability of our solution on both Maximum Cut and TSP, we train our RL agent on small graphs of a fixed size ($n=50$) and test it on larger graphs with sizes up to 300. Table \ref{tab:generalize} summarizes the results with the comparison to S2V-DQN, where the reported values are the approximation ratios averaged over 100 test graphs. As we can see, LS-DQN achieved encouraging and consistent approximation ratios across different test graph sizes on both tasks. LS-DQN generalizes as good as S2V-DQN on Maximum $k$-Cut, and has shown some advantage on TSP tasks. This confirms the generalization of our proposed RL solution framework and the applicability of learned agents across distinct problem instances.

\begin{table}[t]
\caption{Evaluation of generalization of LS-DQN (LS) and comparison with S2V-DQN (S2V). Models are trained on graphs with 50 nodes, and tested on larger sizes up to 300. The average approximation ratio over different testsets is reported.}
\centering
\begin{tabular}{c|c|c|c|c}
\hline 
Test Size & \multicolumn{2}{c}{51-100} & \multicolumn{2}{|c}{101-150}  \\\hline
Algorithm & LS & S2V& LS & S2V\\\hline
MAXCUT& 0.984& 0.988& 0.974 &0.971\\\hline
TSP& 1.034 &1.075& 1.054& 1.089\\\hline
Test Size  & \multicolumn{2}{|c}{151-200} & \multicolumn{2}{|c}{200-300} \\\hline
Algorithm & LS & S2V& LS & S2V \\\hline
MAXCUT&  0.972 &0.975 & 0.978&0.981\\\hline
TSP&  1.088 &1.087& 1.094&1.095\\\hline
\end{tabular}
\normalsize
\label{tab:generalize} 
\vspace{-3mm}
\end{table}

\paragraph{Trade-off between efficiency and accuracy}

\begin{figure}
\centering
\subfigure[Maximum $k$-Cut]{
\label{fig:subfig:a}
\includegraphics[width=0.21\textwidth]{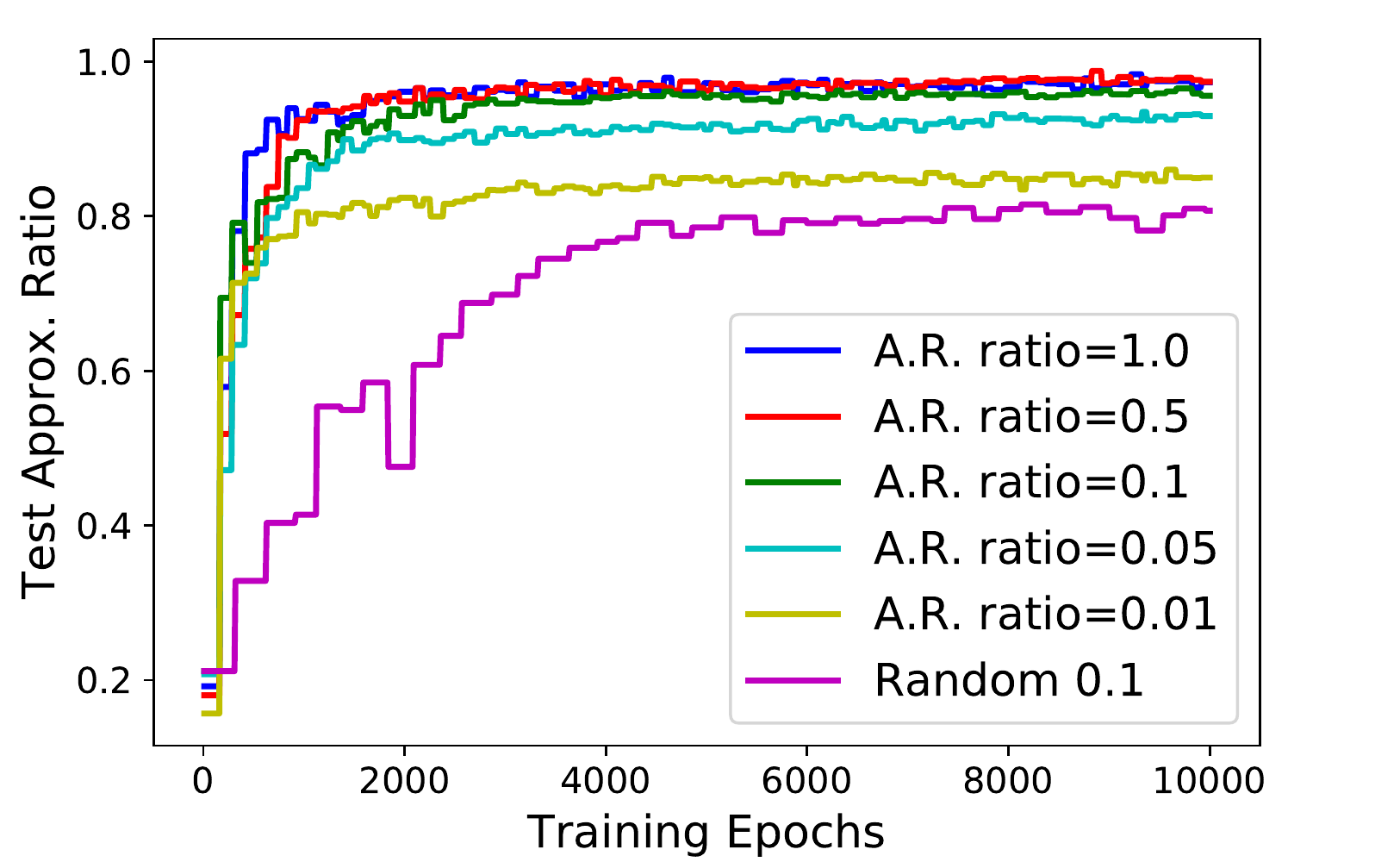}}
\hspace{.2in}
\subfigure[TSP]{
\label{fig:subfig:b}
\includegraphics[width=0.21\textwidth]{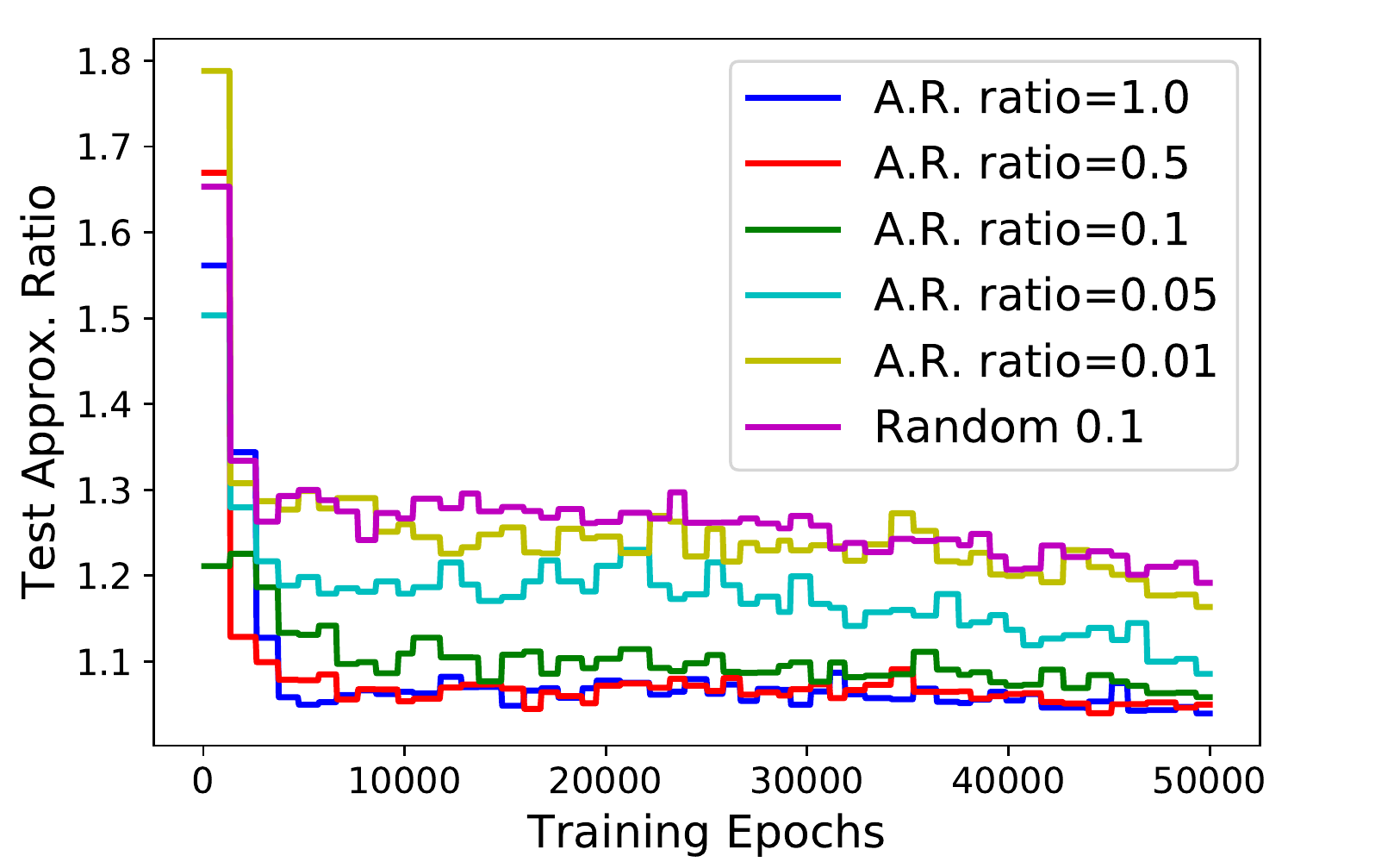}}
\vspace{-3mm}
\caption{Learning curves for Maximum $k$-Cut and TSP with different action reserve ratios. The test performance is evaluated every 200 epochs in Maximum $k$-Cut and every 100 epochs in TSP. 
}
\label{fig:tradeoff} %% label for entire figure
\vspace{-6mm}
\end{figure}

As we have discussed in the End-to-End Training subsection, the deployment of the action-proposal network (AP-net) is essential to enable our method to scale to large graphs. In the AP-net, the action reserve ratio $\epsilon \in (0, 1]$ controls the proportion of actions being evaluated at each state: a large $\epsilon$ allows the agent to evaluate q-values on more state-action pairs, which enhances the quality of planning but also increases the time and space complexity. We investigate the trade-off between efficiency and approximation ratio in our solution. Figure \ref{fig:tradeoff} shows the training curve for Maximum $k$-Cut and TSP under different action reserve ratios. We choose $\epsilon \in \{1.0, 0.5, 0.01, 0.05, 0.01\}$, and a baseline that randomly samples 10\% actions without using the AP-net. As we can find, the performance gap is nearly negligible when $\epsilon \geq 0.1$. When $\epsilon \leq 0.05$, the performance drop becomes sensible, but the approximation ratio when reserving only 1\% actions proposed by the AP-net is still better than randomly selecting 10\% actions. Table \ref{tab:tradeoff} lists the relative running time and memory cost when applying different $\epsilon$. It shows that we can maintain $99\%$ performance in terms of the approximation ratio while only consuming approximately $10\%$ time and space for Maximum $k$-Cut. For TSP, since we adopt an RNN structure in the Q-net, the time and space savings are acceptable but not as significant as in Maximum $k$-Cut.

\begin{table}[t]
\caption{\label{tab:tradeoff kcut}Trade-off between time/space complexity and the Approx. ratio (A.R.) with different action reserve ratios $\epsilon$ in the AP-net. The value shown in the table is scaled according to the result at $\epsilon=1.0$. %For the approximation ratio, the larger the better; for the time and space costs, the lower the better.
}
\centering
\small
\begin{tabular}{c|c|c|c|c|c|c}
\hline
$\epsilon$  & \multicolumn{3}{c}{$k$-Cut} & \multicolumn{3}{|c}{TSP} \\\hline
&  A.R. & Time & Space & A.R. & Time & Space \\\hline
0.5&  1.00& 0.51 & 0.53& 1.00& 0.68& 0.51 \\\hline
0.1&  0.99 & 0.11 & 0.14& 0.98&0.29& 0.12\\\hline
0.05&  0.95&0.06&0.10 & 0.93&0.23&0.09\\\hline
0.01&  0.82 & 0.02&0.06 & 0.80 &0.19&0.06\\\hline
\end{tabular}
\normalsize
\label{tab:tradeoff}
\vspace{-3mm}
\end{table}

\section{Visualization} 
To better understand the RL agent's behavior in solving COPs, we visualize the results obtained from our model.

\paragraph{Trajectory Statistics}
We compare the intra-episode behavior of our proposed method, LS-DQN, with the greedy algorithm equipped with the same action space to illustrate how our LS-DQN agent provides better solutions. We chose greedy algorithm for comparison due to its strong performance in our empirical results reported in the main paper. 

\begin{figure}[t]
\centering
\subfigure[Maximum $k$-Cut]{
\label{fig:subfig:c}
\includegraphics[width=0.21\textwidth]{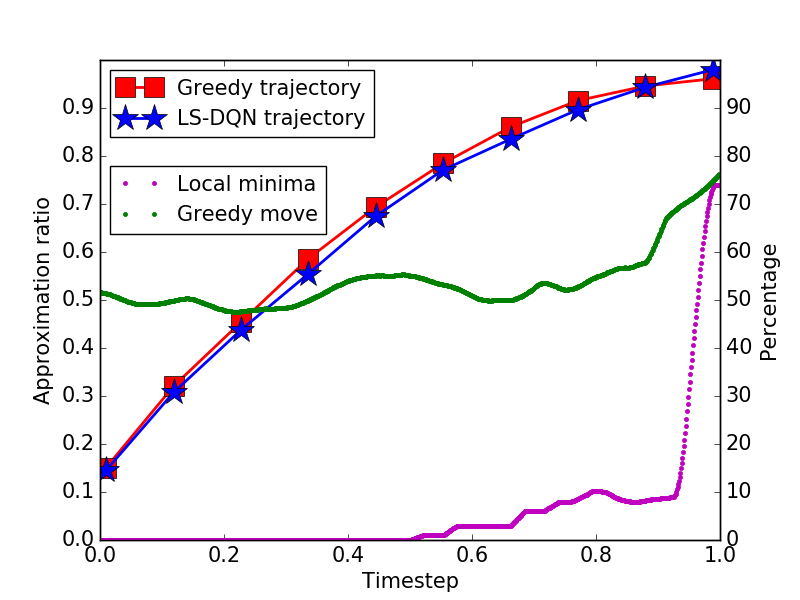}}
\hspace{.2in}
\subfigure[TSP]{
\label{fig:subfig:d}
\includegraphics[width=0.21\textwidth]{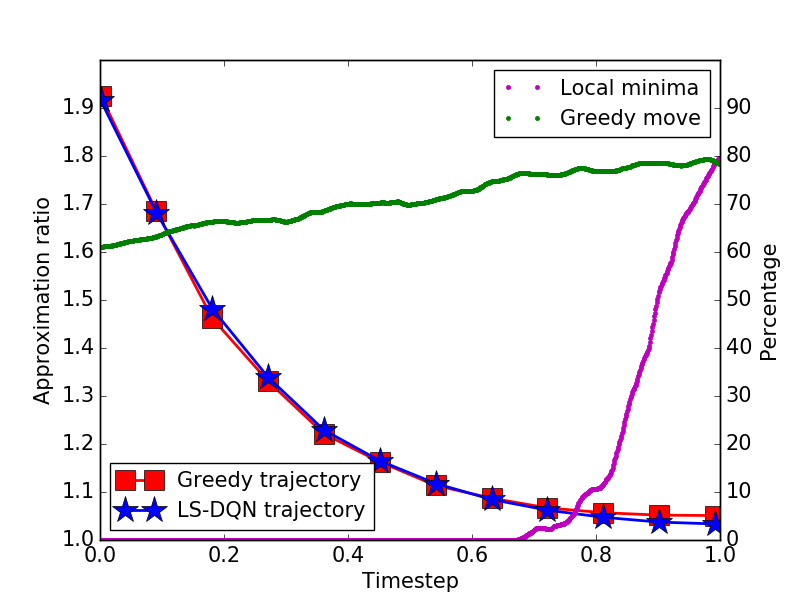}}
\caption{Averaged behavior of LS-DQN over 200 instances of graph size $(k,m)$=50 for Maximum $k$-Cut and $n$=50 for TSP. Timesteps are scaled to the range of [0, 1], where 0 and 1 represent the first state (i.e., a randomly initialized one) and the last state (i.e. where the best solution is obtained within the maximum episode length), respectively. The purple and green dots represent the frequency of LS-DQN stepping into a local minimum and the frequency of taking a greedy move at each time step. The red square and blue star track the improvement of the approximation ratio metric from the greedy algorithm and LS-DQN. The result demonstrates the strength of LS-DQN lies in two aspects: 1. it encourages exploration in the initial stage and exploits greedy move more often at the end; 2. it is able to jump out of the local minimum by taking currently less promising actions.
}
\vspace{-6mm}
\label{fig:episode_stats}
\end{figure}

Figure \ref{fig:episode_stats} shows the traces of the improvement in terms of the approximation ratio metric during test episodes for Maximum $k$-Cut and TSP, respectively. Each episode is terminated at the step where the maximum objective value is reached before hitting the maximum episode length (i.e., 100) or the dummy action was chosen. The traces for LS-DQN (blue star curve) and greedy algorithm (red square curve) are averaged over 200 test instances of size $(k,m)$=50 for Maximum $k$-Cut and $n$=50 for TSP. 
We normalize episode length to [0, 1] in order to calculate the average approximation ratio across episodes with different lengths. In addition, at each timestep, we use the green dots to illustrate how frequently a greedy move (the action with the maximum immediate reward) is taken by LS-DQN, and the purple dots to show how often the agent encounters a local-minimum state (the state where no action with positive immediate reward is available). 

As we can find, in both problems, the trained agent chose greedy actions at a lower frequency at the early stage and it tended to take more greedy actions later in the episode. It indicates that the trained agent learns to explore the solution space at the beginning so that it can benefit more in the future. We should note that in LS-DQN a state encodes a particular solution to the problem instance. When the state (i.e., the current solution) gets closer to the optimal, the LS-DQN agent takes more greedy actions; and that is when the performance of LS-DQN started to surpass the greedy algorithm. The purple dots suggest that although an LS-DQN agent may run into local-minimum states along the way, however, it manages to jump out and land at a better local minimum. For example, in the later stage more and more states have no action with positive immediate reward. A greedy algorithm typically will terminate by then; but the LS-DQN agent managed to take a series of (currently less promising) actions to keep improving the quality of its obtained solution. In our evaluations, the agent secured a good local-minimum state with a high probability of around 80\%. For those cases where an episode ends up in a non-local-minimum state, they are caused by the approximation error of Q-function: the absolute value of the $Q$ estimation shrinks and gets close to zero as the agent approaches the optimal state, therefore the approximation error is more likely to obscure the positive Q-value to a negative one, which results in an early stop. 

\paragraph{Visualization of Learned Strategy}

We illustrate two examples of the search trajectory from a trained LS-DQN agent for Maximum $k$-Cut and TSP, respectively. 

\begin{figure*}
\centering
% \subfigure[Step-0~~O=15.80]{
% \includegraphics[width=0.25\textwidth]{resource/greedy0.png}}
% \hspace{-.2in}
\subfigure[O=13.74]{
\includegraphics[width=0.14\textwidth]{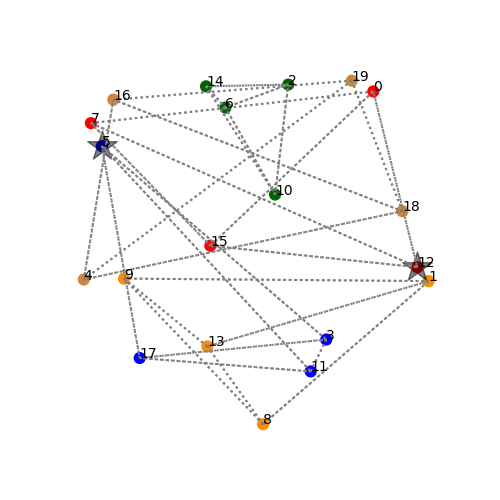}}
\hspace{-.2in}
\subfigure[O=12.18]{
\includegraphics[width=0.14\textwidth]{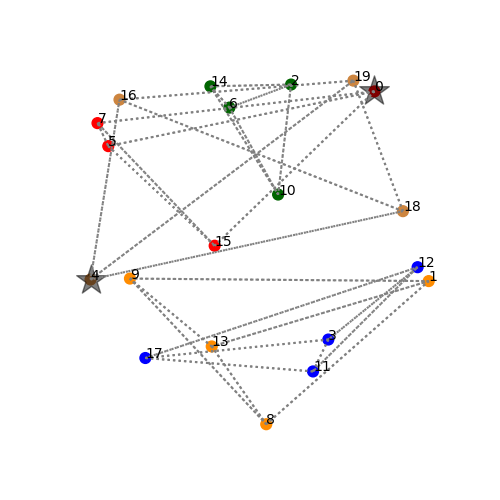}}
\hspace{-.2in}
\subfigure[O=10.53]{
\includegraphics[width=0.14\textwidth]{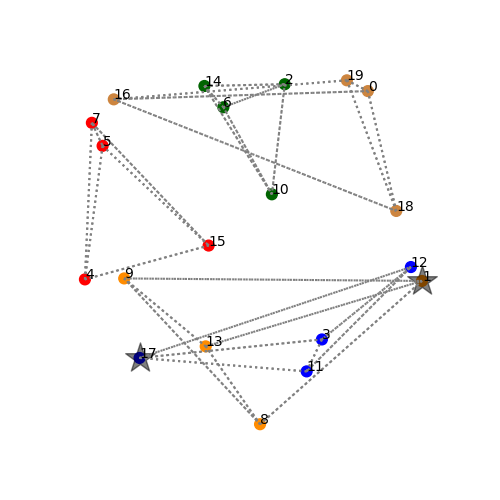}}
\hspace{-.2in}
\subfigure[O=8.89]{
\includegraphics[width=0.14\textwidth]{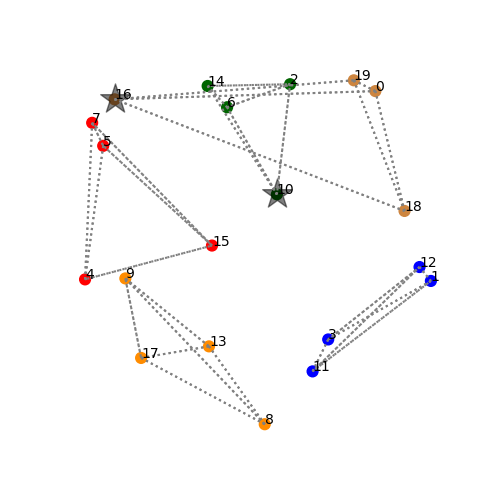}}
\hspace{-.2in}
\subfigure[O=8.07]{
\includegraphics[width=0.14\textwidth]{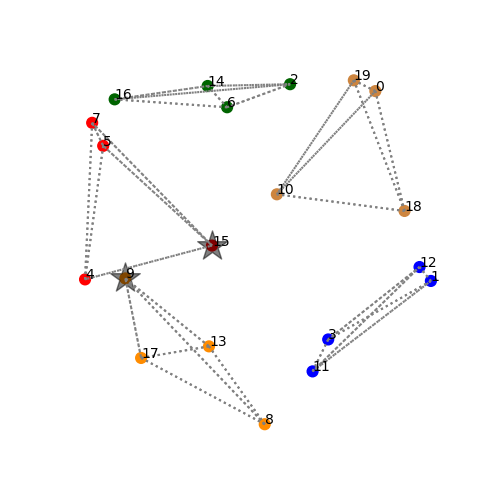}}
\hspace{-.2in}
\subfigure[O=8.04]{
\includegraphics[width=0.14\textwidth]{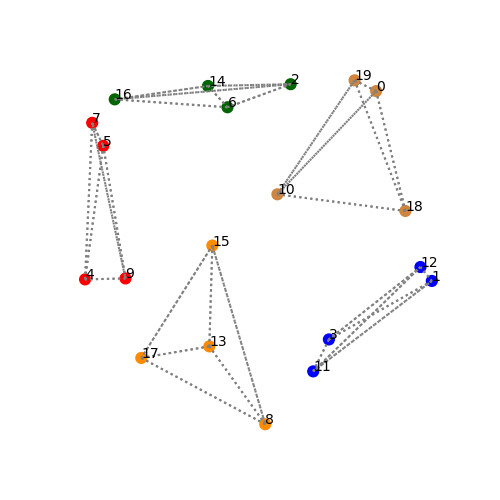}}
\hspace{-.2in}
\caption{Sample episode of greedy algorithm for Maximum $k$-Cut}
\label{fig:kcut-greedy} %% label for entire figure
\end{figure*}

\begin{figure*}
\centering
% \subfigure[Step-0~~O=15.80]{
% \includegraphics[width=0.25\textwidth]{resource/dqn0.png}}
% \hspace{-.2in}
\subfigure[O=13.74]{
\includegraphics[width=0.12\textwidth]{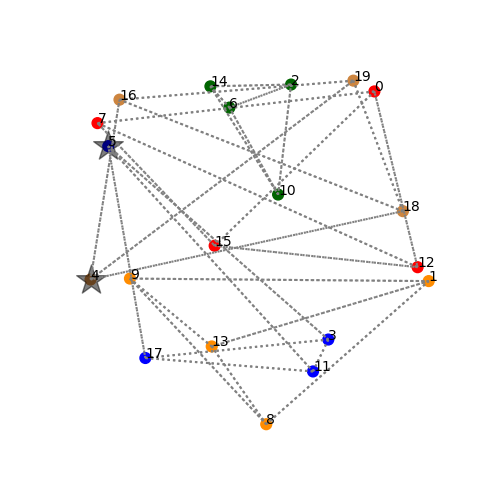}}
\hspace{-.2in}
\subfigure[O=12.25]{
\includegraphics[width=0.12\textwidth]{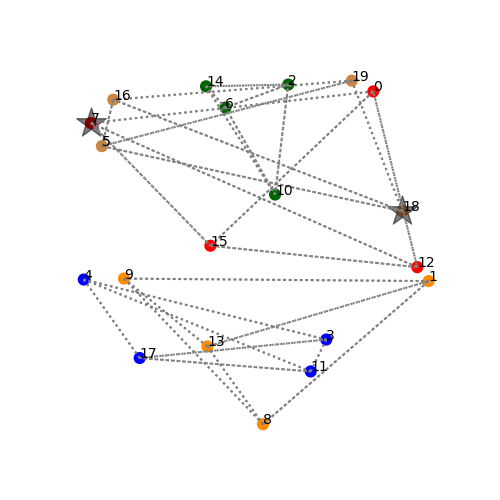}}
\hspace{-.2in}
\subfigure[O=10.38]{
\includegraphics[width=0.12\textwidth]{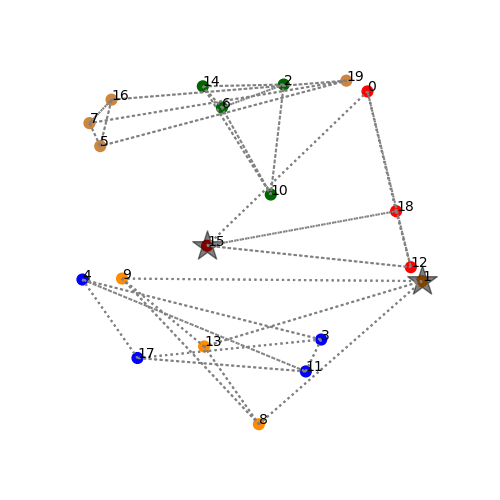}}
\hspace{-.2in}
\subfigure[O=9.19]{
\includegraphics[width=0.12\textwidth]{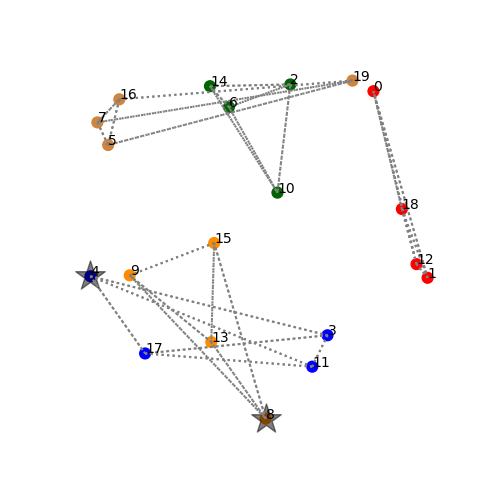}}
\hspace{-.2in}
\subfigure[O=8.05]{
\includegraphics[width=0.12\textwidth]{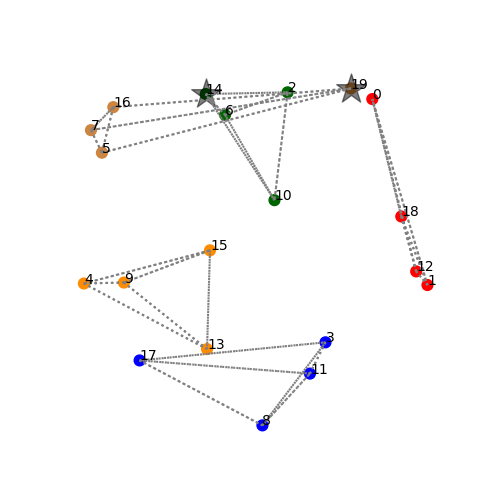}}
\hspace{-.2in}
\subfigure[O=7.38]{
\includegraphics[width=0.12\textwidth]{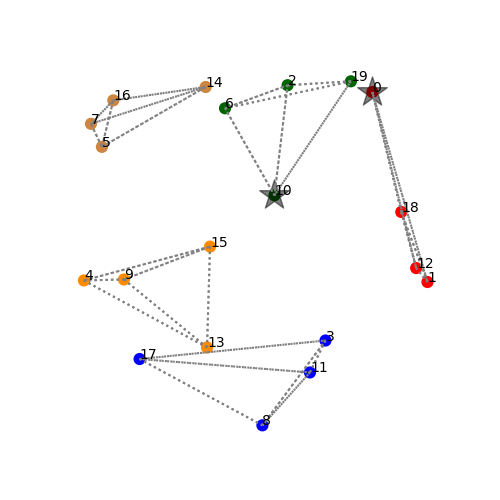}}
\hspace{-.2in}
\subfigure[O=6.54]{
\includegraphics[width=0.12\textwidth]{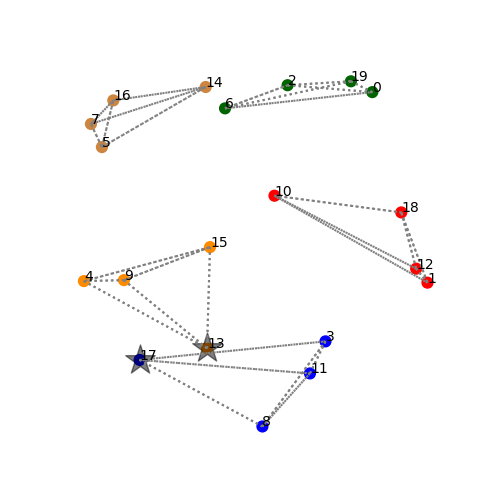}}
\hspace{-.2in}
\subfigure[O=6.33]{
\includegraphics[width=0.12\textwidth]{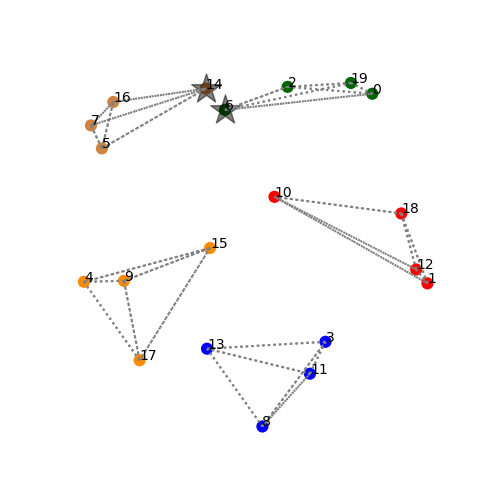}}
\hspace{-.2in}
\subfigure[O=6.19]{
\includegraphics[width=0.12\textwidth]{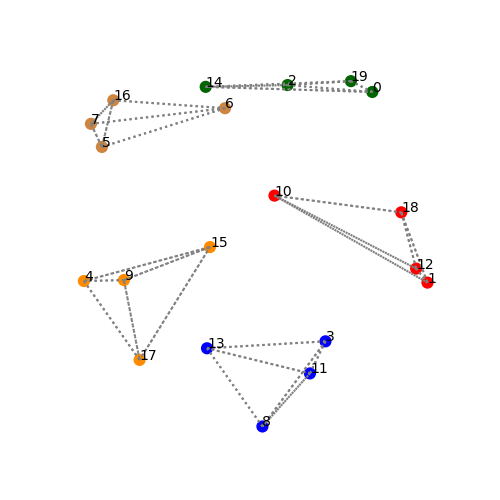}}
\hspace{-.2in}
\caption{Sample episode of LS-DQN for $k$-Cut}
\label{fig:kcut-dqn} %% label for entire figure
\end{figure*}

Figure \ref{fig:kcut-greedy} and \ref{fig:kcut-dqn} show the comparison between greedy algorithm and our LS-DQN on a Maximum $k$-Cut instance of size $(k, m)$=(5,4). For the ease of presentation, we plot the complementary graph instead of the entire cut and revise the objective value $O$ as the sum of edge-weights within each cluster. The graph cut result is visualized by 5 different colors on vertices and the chosen swapping action at each step is marked by stars. In Figure \ref{fig:kcut-greedy}, the greedy algorithm starts from an initial solution with $O=13.74$ and stops at $O=8.04$ in five steps, as no further swapping actions can immediately improve the current solution. Clearly this greedy strategy is trapped by the locally stable clusters (e.g., the green and blue cluster in Figure \ref{fig:kcut-greedy} at step 6), which look optimal in the local region but are not necessarily globally optimal. As a contrast to the greedy heuristic, Figure \ref{fig:kcut-dqn} shows how LS-DQN arrives at a better solution from the same initial state. It chose not to take any greedy action in the first 5 steps to avoid the formation of locally stable clusters and then started to exploit greedy actions to refine the cut locally from the 6th step. As a result, it yielded a better solution at the end.

\begin{figure*}
\centering
% \subfigure[Step-0~~O=15.80]{
% \includegraphics[width=0.25\textwidth]{resource/tspgreedy0.png}}
% \hspace{-.2in}
\subfigure[O=8.63]{
\includegraphics[width=0.14\textwidth]{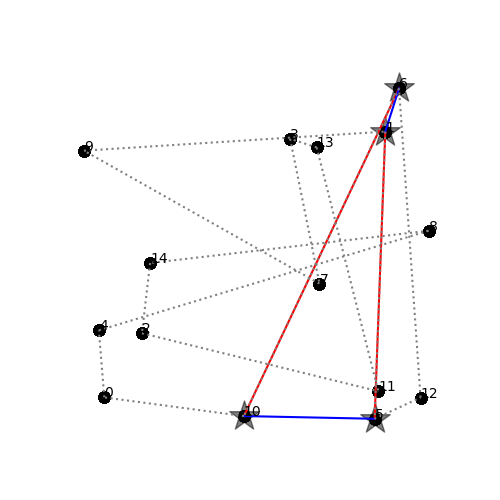}}
\hspace{-.2in}
\subfigure[O=7.34]{
\includegraphics[width=0.14\textwidth]{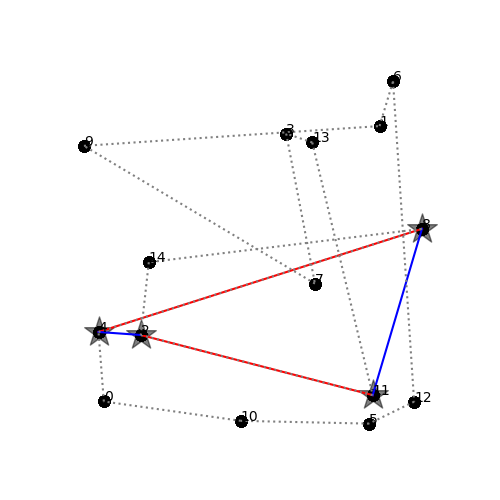}}
\hspace{-.2in}
\subfigure[O=6.31]{
\includegraphics[width=0.14\textwidth]{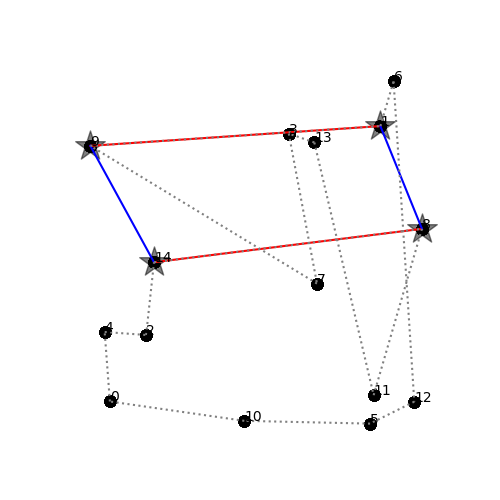}}
\hspace{-.2in}
\subfigure[O=5.36]{
\includegraphics[width=0.14\textwidth]{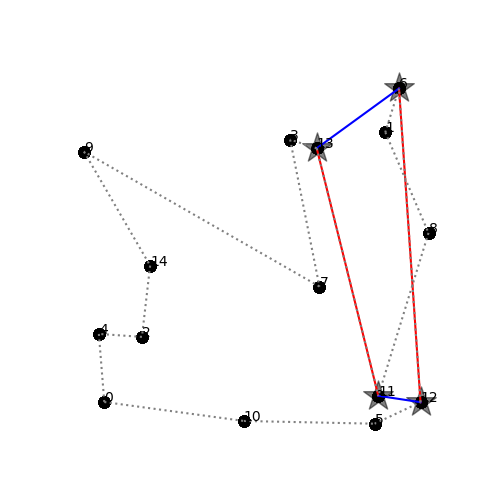}}
\hspace{-.2in}
\subfigure[O=4.22]{
\includegraphics[width=0.14\textwidth]{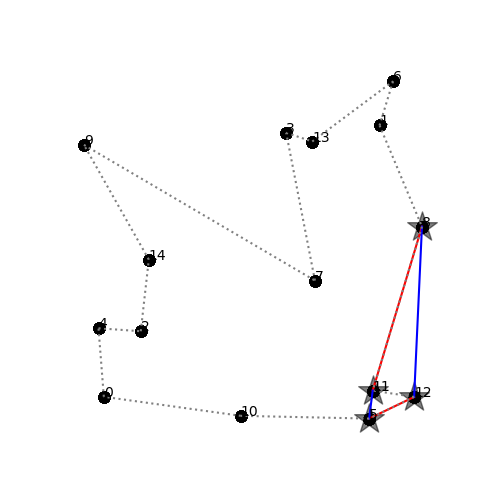}}
\hspace{-.2in}
\subfigure[O=4.16]{
\includegraphics[width=0.14\textwidth]{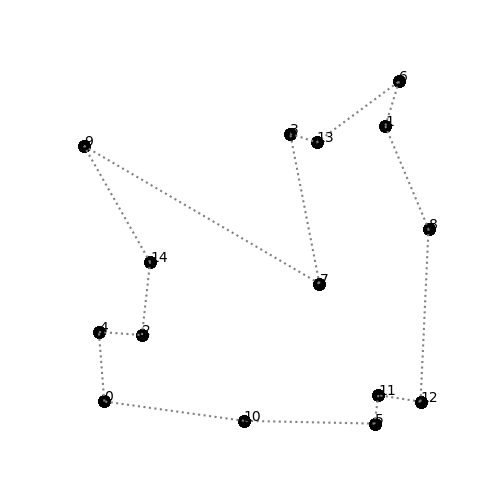}}
\hspace{-.2in}
\caption{Sample episode of greedy heuristic for TSP}
\label{fig:tsp-greedy} %% label for entire figure
\end{figure*}

\begin{figure*}
\centering
% \subfigure[Step-0~~O=15.80]{
% \includegraphics[width=0.25\textwidth]{resource/tspdqn0.png}}
% \hspace{-.2in}
\subfigure[O=8.63]{
\includegraphics[width=0.12\textwidth]{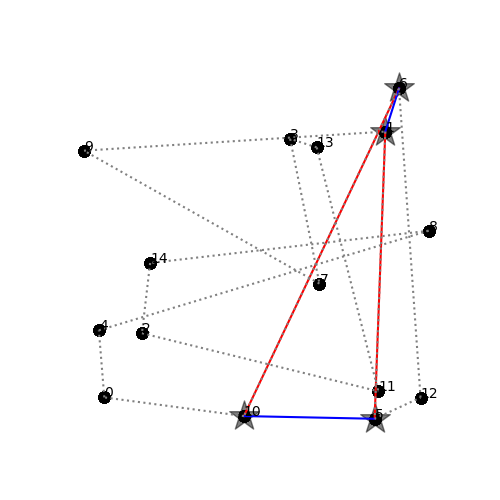}}
\hspace{-.2in}
\subfigure[O=7.34]{
\includegraphics[width=0.12\textwidth]{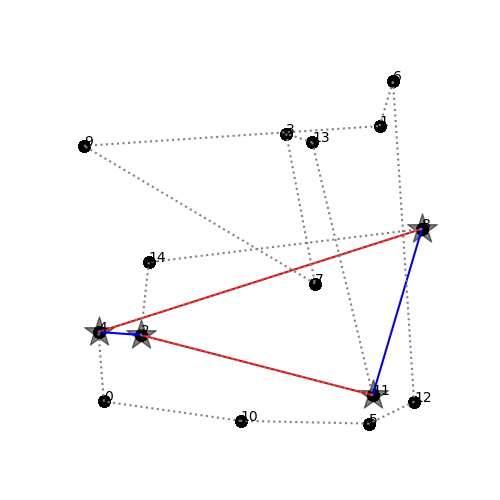}}
\hspace{-.2in}
\subfigure[O=6.31]{
\includegraphics[width=0.12\textwidth]{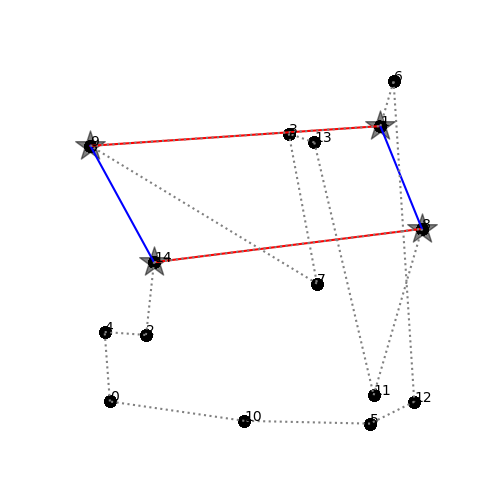}}
\hspace{-.2in}
\subfigure[O=5.36]{
\includegraphics[width=0.12\textwidth]{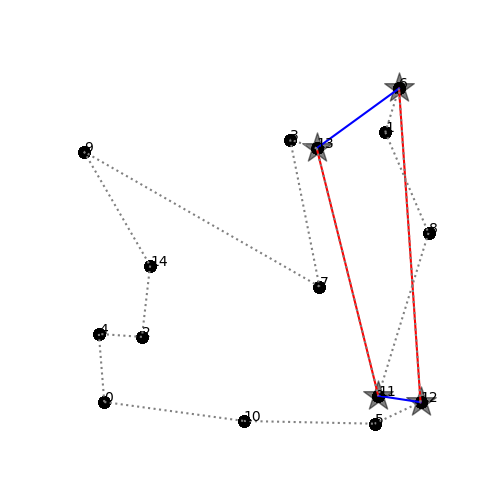}}
\hspace{-.2in}
\subfigure[O=4.22]{
\includegraphics[width=0.12\textwidth]{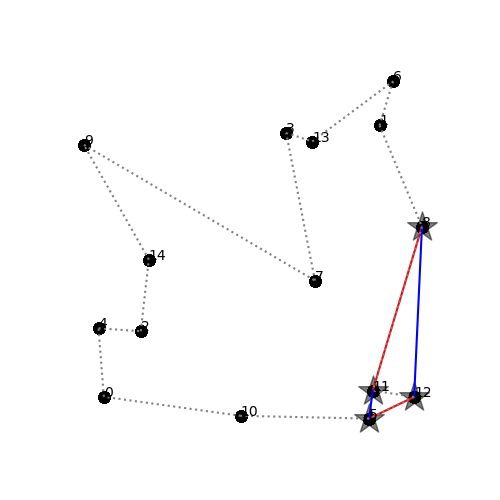}}
\hspace{-.2in}
\subfigure[O=4.16]{
\includegraphics[width=0.12\textwidth]{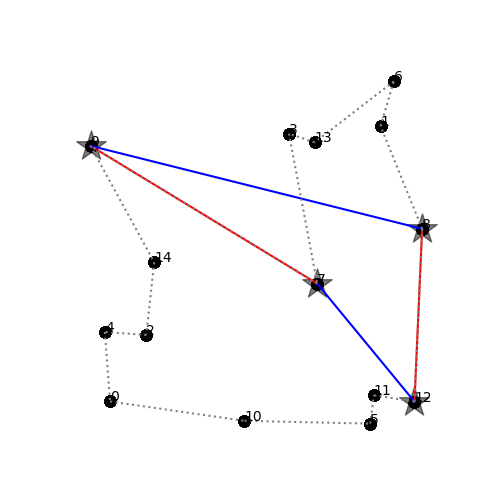}}
\hspace{-.2in}
\subfigure[O=4.35]{
\includegraphics[width=0.12\textwidth]{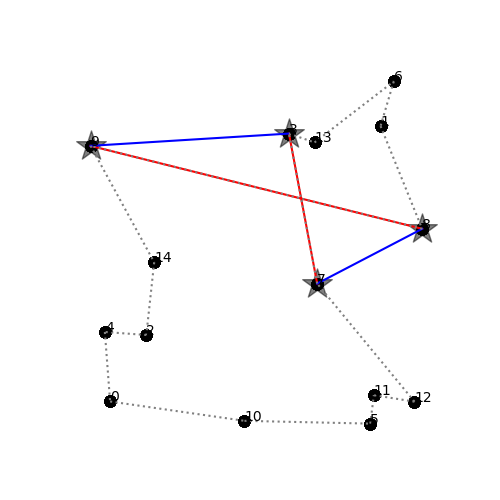}}
\hspace{-.2in}
\subfigure[O=3.88]{
\includegraphics[width=0.12\textwidth]{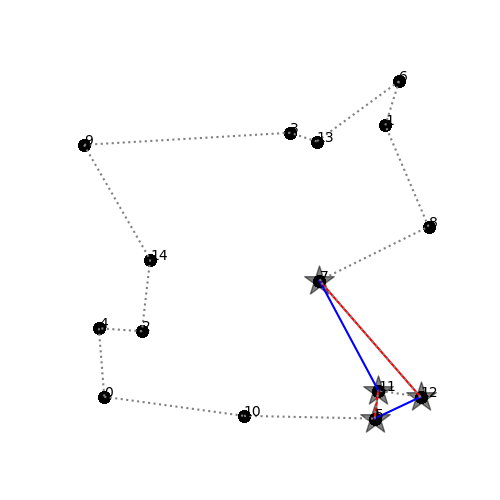}}
\hspace{-.2in}
\subfigure[O=3.86]{
\includegraphics[width=0.12\textwidth]{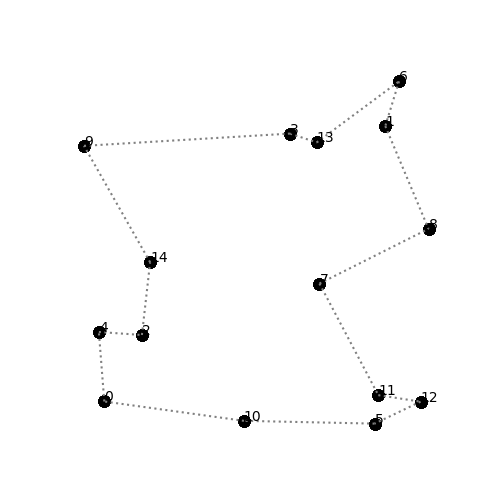}}
\hspace{-.2in}
\caption{Sample episode of LS-DQN for TSP}
\label{fig:tsp-dqn} %% label for entire figure
\end{figure*}

Figure \ref{fig:tsp-greedy} and \ref{fig:tsp-dqn} report the same comparison on TSP of size $n$=15. The TSP tour is visualized by the dashed lines between vertices, the swapping action is marked by stars on vertices, and the solid lines colored in red and blue suggest the two red lines will be replaced by the two blue lines after the swapping action. Figure \ref{fig:tsp-greedy} shows the greedy algorithm (i.e., 2-opt) that starts from a tour with $O=8.63$ and stops at $O=4.16$ in five steps, yielding a locally optimal solution. Interestingly, Figure \ref{fig:tsp-dqn} shows how LS-DQN got a better solution by a different search trajectory. Starting from the same solution as in Figure \ref{fig:tsp-greedy}, the LS-DQN agent behaved exactly the same as the greedy heuristic in the first 5 steps and arrives at the same local minimum. However, it managed to take an aggressive move which increases the tour length from $O=4.16$ to $O=4.35$. After taking this seemingly bad action, the agent made two consecutive greedy actions to further reduce the tour length from $O=4.35$ to $O=3.86$. This example demonstrates LS-DQN's ability to jump out of local minima by making farsighted decisions. 

\section{Conclusions and Future Work}
In this paper, we introduced a general end-to-end RL framework for solving combinatorial optimization problems on graphs. The key idea behind our design is to view a solution to a problem instance as state and reversible perturbation to this solution as action. We introduce graph neural networks to extract latent representations of graphs for state-action encoding, and  apply deep Q-learning to obtain a policy that gradually improves the solution. We instantiated the meta-algorithm for Maximum $k$-Cut and TSP, where extensive experiment results demonstrate the solution's competitive performance and generalization across problem instances. The major obstacle that prevents its application to large graphs is the formidable size of swapping action space, which has been successfully addressed by introducing an auxiliary action-proposal network. One important direction of this work is the refinement of the state representation, since our current design does not fully consider the combinatorial structure of the solution: we use one-hot labels on vertices to represent a graph cut and a sequence to represent a tour in TSP. Although they work well empirically, to design a network architecture that encodes various nontrivial combinatorial structures remains a challenging yet promising direction.
% The permutation/rotation invariant issue raises some concern: solutions that are equivalent under a permutation or rotation of vertex labels will lead to a totally different representation. This costs an unnecessarily large input space and burden on the capacity of the neural network. Several existing works also mentioned this issue \cite{zaheer2017deep,rezatofighi2018deep,meltzer2019pinet,wagstaff2019limitations}, but did not provide an effective solution yet.
% %they mainly focused on the representation of sets, which is not sophisticated enough for most combinatorial problems: in the case of Maximum $k$-Cut, the solution is a nested set and in TSP the solution is a cycle). 
% To design a network architecture that encodes various nontrivial combinatorial structures remains a challenging yet promising direction.

% \begingroup
% \let\clearpage\relax 
% \onecolumn 

\begin{table*}[p]
\centering
\caption{Approximation ratio comparison for Maximum $k$-Cut ($k=2$). Larger is better.}
\begin{tabular}{c|c|c|c|c|c|c}
\hline
Train/Test Size & LS-DQN & S2V-DQN & ECO-DQN & SDP & GP & Greedy \\\hline
$n=20$ & $0.98^{+0.02}_{-0.02}$& $0.98^{+0.01}_{-0.01}$& $0.99^{+0.01}_{-0.02}$& $0.98$& $0.96^{+0.03}_{-0.03}$& $0.98^{+0.02}_{-0.02}$\\\hline
$n=50$ & $0.98^{+0.02}_{-0.02}$& $0.97^{+0.02}_{-0.02}$& $0.98^{+0.02}_{-0.02}$& $0.97$& $0.96^{+0.03}_{-0.03}$& $0.97^{+0.02}_{-0.02}$\\\hline
$n=100$ & $0.97^{+0.01}_{-0.01}$& $0.94^{+0.02}_{-0.02}$& $0.97^{+0.02}_{-0.02}$& $0.95$& $0.92^{+0.03}_{-0.03}$& $0.94^{+0.02}_{-0.02}$\\\hline
$n=200$ & $0.95^{+0.01}_{-0.01}$& $0.94^{+0.02}_{-0.02}$& $0.96^{+0.01}_{-0.01}$& $0.89$& $0.85^{+0.02}_{-0.02}$& $0.93^{+0.02}_{-0.02}$\\\hline
$(k, m)=(5, 6)$& $0.98^{+0.02}_{-0.02}$& -&-&  $0.92$& $0.67^{+0.05}_{-0.05}$& $0.98^{+0.02}_{-0.02}$\\\hline
$(k, m)=(10, 10)$& $0.97^{+0.02}_{-0.02}$& -&-& $0.89$& $0.42^{+0.05}_{-0.05}$& $0.96^{+0.02}_{-0.02}$\\\hline
$(k, m)=(10, 20)$& $0.95^{+0.02}_{-0.02}$& -&-& $0.86$& $0.34^{+0.04}_{-0.04}$& $0.94^{+0.02}_{-0.02}$\\\hline
\end{tabular}
\label{tab:kcut}
\vspace{-3mm}
\end{table*}
% \FloatBarrier

% \begin{table*}[p]
% \caption{Evaluation of generalization of LS-DQN (LS) and comparison with S2V-DQN (S2V). Models are trained on graphs with 50 nodes, and tested on larger sizes up to 300. The average approximation ratio over different testsets is reported.}
% \centering
% \begin{tabular}{c|c|c|c|c|c|c|c|c}
% \hline 
% Test Size & \multicolumn{2}{c}{51-100} & \multicolumn{2}{|c}{101-150}  & \multicolumn{2}{|c}{151-200} & \multicolumn{2}{|c}{200-300} \\\hline
% Algorithm & LS & S2V& LS & S2V& LS & S2V& LS & S2V \\\hline
% MAXCUT& 0.984& 0.988& 0.974 &0.971& 0.972 &0.975 & 0.978&0.981\\\hline
% TSP& 1.034 &1.075& 1.054& 1.089& 1.088 &1.087& 1.094&1.095\\\hline
% \end{tabular}
% \normalsize
% \label{tab:generalize} 
% \vspace{-3mm}
% \end{table*}

\begin{table*}[p]
\centering
\caption{Approximation ratio comparison for TSP. Smaller is better.}
\begin{tabular}{c|c|c|c|c|c}
\hline
Train Size & Test Size & LS-DQN  & S2V-DQN & Farthest & 2-opt \\\hline
$n=50$&51-100& $1.04^{+0.01}_{-0.01}$& $1.05^{+0.01}_{-0.01}$& $1.07^{+0.01}_{-0.01}$& $1.07^{+0.01}_{-0.01}$\\\hline
$n=100$&101-150& $1.05^{+0.01}_{-0.01}$& $1.05^{+0.01}_{-0.01}$& $1.08^{+0.01}_{-0.01}$& $1.09^{+0.01}_{-0.01}$\\\hline
$n=150$&151-200& $1.06^{+0.01}_{-0.01}$& $1.07^{+0.01}_{-0.01}$& $1.08^{+0.01}_{-0.01}$& $1.09^{+0.01}_{-0.01}$\\\hline
\end{tabular}
\label{tab:tsp}
\vspace{-3mm}
\end{table*}

% \begin{table*}[p]
% \caption{\label{tab:tradeoff kcut}Trade-off between time/space complexity and the Approx. ratio (A.R.) with different action reserve ratios $\epsilon$ in the AP-net. The value shown in the table is scaled according to the result at $\epsilon=1.0$. %For the approximation ratio, the larger the better; for the time and space costs, the lower the better.
% }
% \centering
% \small
% \begin{tabular}{c|c|c|c|c|c|c|c|c|c|c|c|c}
% \hline
% $\epsilon$  & \multicolumn{3}{c}{0.5} & \multicolumn{3}{|c}{0.1} & \multicolumn{3}{|c}{0.05} & \multicolumn{3}{|c}{0.01} \\\hline
% &  A.R. & Time & Space & A.R. & Time & Space & A.R. & Time & Space & A.R. & Time & Space  \\\hline
% $k$-Cut&  1.00& 0.51 & 0.53& 0.99& 0.11&0.14& 0.95&0.06&0.10& 0.82 &0.02&0.06\\\hline
% TSP&  1.00 & 0.68&0.51& 0.98&0.29&0.12& 0.93&0.23&0.09& 0.80 &0.19&0.06\\\hline
% \end{tabular}
% \normalsize
% \label{tab:tradeoff}
% \vspace{-3mm}
% \end{table*}

% \endgroup